\documentclass[conference]{IEEEtran}

\ifCLASSINFOpdf

\else

\fi

\usepackage{url}
\usepackage{graphicx}
\usepackage{xcolor}
\graphicspath{{figures/}}
\usepackage{relsize}

\usepackage{listings}

\usepackage{float}
\usepackage{caption}
\usepackage{subcaption}
\usepackage{color}
\usepackage{comment}
\usepackage{pgfplots}
\pgfplotsset{compat=1.9}
\usepackage{booktabs}
\usepackage{multirow}
\usepackage{textpos}
\usepackage{booktabs}
\usepackage{url}
\usepackage{bm}
\usepackage{mathtools}
\usepackage{multicol}
\usepackage{lipsum}
\usepackage{mwe}
\usepackage{amsmath}
\usepackage{commath}
\usepackage{bm}

\usepackage{soul}
\usepackage{siunitx}
\usepackage[numbers]{natbib}
\bibpunct{[}{]}{,}{n}{,}{;}
\usepackage{pdfpages}
\pdfoutput=1

\begin{document}

\title{
{\footnotesize Preprint identical to the final version of 2019 IEEE 15th International Conference on Intelligent Computer Communication and Processing (ICCP 2019).} \newline
\newline
Generating Data using Monte Carlo Dropout}

\author{
\begin{tabular}[t]{c@{\extracolsep{1em}}c@{\extracolsep{1em}}c} 
Kristian Miok  & Dong Nguyen-Doan & Daniela Zaharie \\
West University of Timisoara & West University of Timisoara & West University of Timisoara \\ 
Computer Science Department & Computer Science Department  & Computer Science Department \\
Romania & Romania &  Romania\\
Email: kristian.miok@e-uvt.ro & Email: dong.nguyen10@e-uvt.ro & Email: daniela.zaharie@e-uvt.ro\\
\\
  & Marko Robnik-\v Sikonja \\
  & University of Ljubljana \\ 
  & Faculty of Computer and Information Science \\
  & Slovenia \\
  & Email: marko.robnik@fri.uni-lj.si
\end{tabular}
}



\maketitle

\begin{abstract}
For many analytical problems the challenge is to handle huge amounts of available data. However, there are data science application areas where collecting information is difficult and costly, e.g., in the study of geological phenomena, rare diseases, faults in complex systems, insurance frauds, etc. In many such cases, generators of synthetic data with the same statistical and predictive properties as the actual data allow efficient simulations and development of tools and applications. In this work, we propose the incorporation of Monte Carlo Dropout method within Autoencoder (MCD-AE) and Variational Autoencoder (MCD-VAE) as efficient generators of synthetic data sets. As the Variational Autoencoder (VAE) is one of the most popular generator techniques, we explore its similarities and differences to the proposed methods. We compare the generated data sets with the original data based on statistical properties, structural similarity, and predictive similarity. The results obtained show a strong similarity between the results of VAE, MCD-VAE and MCD-AE; however, the proposed methods are faster and can generate values similar to specific selected initial instances. 



\end{abstract}

\IEEEpeerreviewmaketitle

\section{Introduction}

We live in times of big data; yet, there are many application areas that lack sufficient data for analyses, simulations, and development of analytical approaches. For example, many studies within bio-medical domain require strict and expensive experimental conditions and can produce only small samples within the allocated budget. Similar examples are domains for which data is difficult to obtain, such are rare diseases, private records, or rare grammatical structures \cite{robnik2015data}. Thus, there is a need for machine learning methods that can generate new data preserving the statistical and predictive characteristics of the original data set.

Since its introduction by \citet{diederik2014auto}, Variational autoencoders (VAE) become one of the most used unsupervised learning methods within the family of autoencoder (AE) techniques \cite{doersch2016tutorial}. They are used in various problems: predicting dense trajectories of pixels in computer vision \cite{walker2016uncertain}, anomaly detection \cite{an2015variational},  and conversion of molecular discrete representations to and from multidimensional continuous representations \cite{gomez2018automatic}. A short description of VAEs is provided in Section 3. Our interest in VAEs is due to their ability to generate new data \cite{im2016generating,hu2017toward}.


The main goal of this work is to introduce Monte Carlo dropout into (variational) autoencoder-based data generating methods that can provide comparable results to existing VAE generators in a shorter time. To show favorable properties of the new generators, we conduct comparisons among three groups of data sets: 

\begin{enumerate}
    \item original data sets,
    \item data sets produced by the VAE generator,
    \item data sets generated using the newly introduced MCD-VAE and MCD-AE approaches.
\end{enumerate}
We compare statistics of individual attributes in each of the data sets, structures of the data sets as determined by clustering algorithms, predictive performance of machine learning algorithms trained and tested on data sets from each group, and times required for generation of new instances.

The outline of the paper is as follows. In Section 2, we shortly discuss related work. In Section 3, we introduce the methodology and architecture of our methods. Section 4 describes how the VAE, MCD-VAE and MCD-AE generators were compared followed by the results obtained in Section 5. We compare the computational performance of the three generators in Section 6 and derive conclusions in Section 7. 

\section{Related Work}

Methods that learn the distribution from existing data in order to generate new instances are of recent interest to scientific community. Till recently, generative methods were based on models that provide a parametric specification of a probability distribution function and models that can estimate kernel density \cite{goodfellow2014generative}. For example, \cite{li2006using} and \citet{yang2011novel} used kernel density estimation to generate new virtual instances. However, those methods work only for data sets with low dimensionality. An interesting method that generates new records using an evolutionary algorithm was proposed in \cite{meraviglia2006gend}. This method does not take dependencies between attributes into account. The generator based on Radial Basis Function (RBF) networks \cite{robnik2015data} corrects this shortcoming but is less suitable for really high dimensional data sets (such as images and text). Two popular generators for images are VAEs \cite{gregor2015draw} and Generative Adversarial Networks (GAN) \cite{goodfellow2014generative}. Interesting combinations of those two methods were proposed by \citet{larsen2015autoencoding} and \citet{rosca2017variational} suggesting that a GAN discriminator can be used in place of a VAE’s decoder.

As the GAN generated data that can be very different from the original data set its outputs cannot be used to simulate the original data. On the other hand, the shortcoming of VAE is that the newly generated values strongly depend on the distribution of the whole training set. Hence, in case we want to generate instances similar to specific instances, e.g., outliers, this is impossible. The proposed method addresses the mentioned shortcomings of VAEs and improves upon it in terms of flexibility of the generated instances and speed of generation. 

\section{Methods}
\label{sec:methods}
We first present the background information on AE, VAE and Monte Carlo Dropout method and then explain how we can harness the power of both to produce  flexible and efficient data generators.
Finally, we visually demonstrate the differences between different generators on a digit recognition data set.

\subsection{(Variational) Autoencoders}
\begin{figure}[h]
  \centering
    \includegraphics[width=\linewidth, height=4cm]{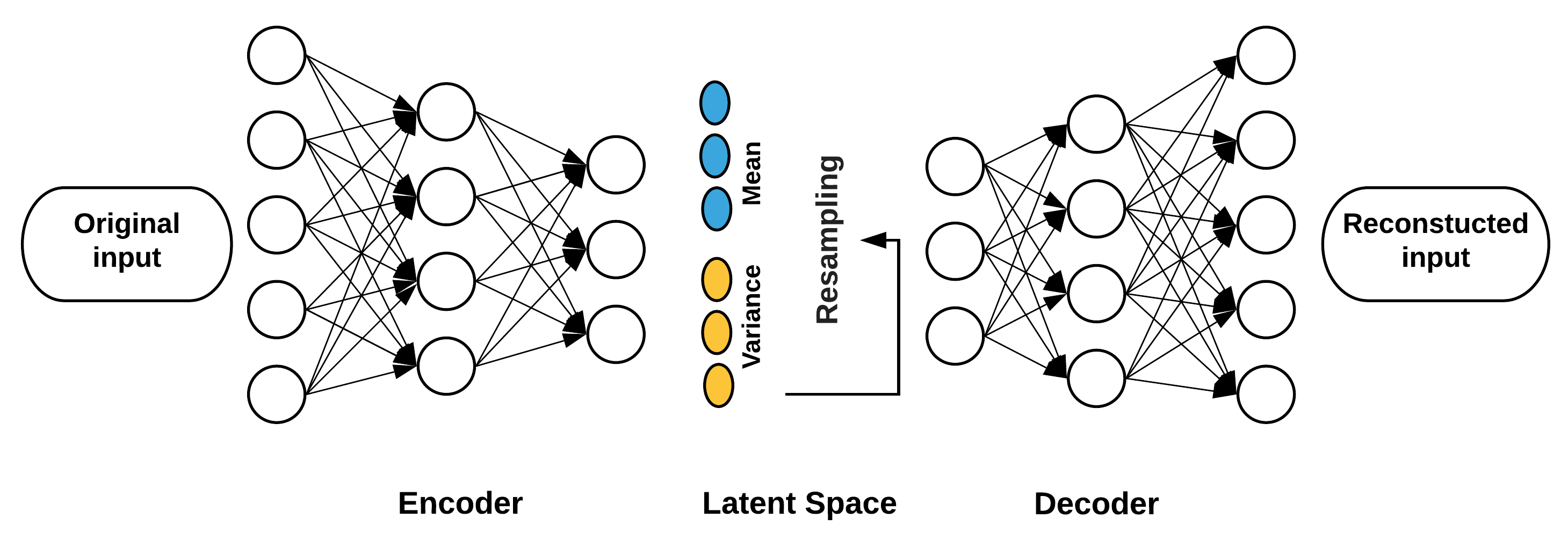}
        \caption{Variational Autoencoder Diagram.
        }
        \label{fig:int}
\end{figure}

A typical AE is made of two neural networks called an encoder and a decoder. The encoder compresses the data into an internal representation and the decoder tries to decompress from this compressed representation (or latent vector) back into the original data using a reconstruction loss function \cite{baldi2012autoencoders}. VAEs inherit the architecture of classical AEs introduced by Rumelhart at al. \cite{rumelhart1985learning}; however, their learning process uses the data to explicitly estimate \emph{the distribution} from which the latent space is sampled \cite{doersch2016tutorial}. Hence, VAEs store the latent variables in the form of  probability distributions. As depicted in Fig. \ref{fig:int}, VAEs resample latent values $z$ from the generated distribution that are further transformed using the decoder network. 
From the Bayesian perspective the encoder is doing an approximation of the posterior distribution $p(z|x)$: 

\begin{equation*}
    p(z|x)=\frac{p(x|z)p(z)}{p(x)},
\end{equation*}
where $z$ denotes the hidden variable values and $x$ the input data. As this distribution usually does not have analytical closed form solution, we have to approximate it. In order to avoid computationally expensive sampling procedure like Markov Chain Monte Carlo (MCMC) sampling, the Variational Inference (VI) method is applied. The VI method \cite{jordan1999introduction} samples from the distribution for which the Kullback-Leibler divergence to the posterior distribution is minimal.     

\subsection{Monte Carlo Dropout Method}

Deep learning is the state-of-the-art approach for many problems where machine learning is applied. However, standard deep neural networks do not provide information on reliability of predictions. Bayesian neural networks (BNN) can overcome this issue by probabilistic interpretation of model parameters. Apart from prediction uncertainty estimation, BNNs offer robustness to overfitting and can be efficiently trained even on  small data sets \cite{kucukelbir2017automatic}. While there exist several BNN variants and implementations, our work is based on Monte Carlo Dropout (MCD) method proposed by \citet{gal2016dropout}. The idea of this approach is to capture prediction uncertainty using the dropout as a regularization technique. Authors prove that the use of dropout in NNs can be seen as a Bayesian approximation of the Gaussian process probabilistic models.
Generating new values can be seen as the uncertainty estimation process of predicting the original instance for which generation is done \cite{miok2018estimation}. The generated values shall reflect the distributional properties of the original instances. 

The bias in the prediction accuracy can come from different sources. Based on where uncertainty is coming from, we distinguish: model uncertainty, data uncertainty, and distributional uncertainty. Model uncertainty describes how well the model fits the data and it can be reduced using larger training set. The data uncertainty is caused by the nature of the data set used and is irreducible by current techniques. Distributional uncertainty arises from the distributional incompatibility between the training and testing data sets. In case of the Bayesian inference, the overall uncertainty is captured with the data and model uncertainty \cite{malinin2018predictive}. The prediction uncertainties within the Bayesian framework can be summarized with the posterior predictive distribution (PPD) \cite{myshkov2016posterior}. Once the posterior distribution is estimated, the PPD can be calculated using the formula: 
    $$ p (y^*|x^*,X,Y) = \int p \big (y^*|f^\omega(x^*)\big) \; p(\omega|X,Y) d\omega $$
where the $p \big (y^*|f^\omega(x^*)\big)$ is the likelihood function that contains the data uncertainty while the $p(\omega|X,Y)$ is the posterior distribution of the model parameters $\omega$ presenting uncertainty of the model.  

The idea of MCD method is to replace the complex Bayesian process of seizing those uncertainties during the regularization using dropout. Practically, the dropout is equivalent to several forward passes through the network and recalculation of the results. At each backward pass, the model ends-up with new optimization results of the model weights. Keeping all this information, the method  mimics the Bayesian inference and is equivalent to the Bayesian posterior distribution estimation \cite{gal2016uncertainty}.

\begin{figure}[hbt]
  \centering
    \includegraphics[width=\linewidth, height=4cm]{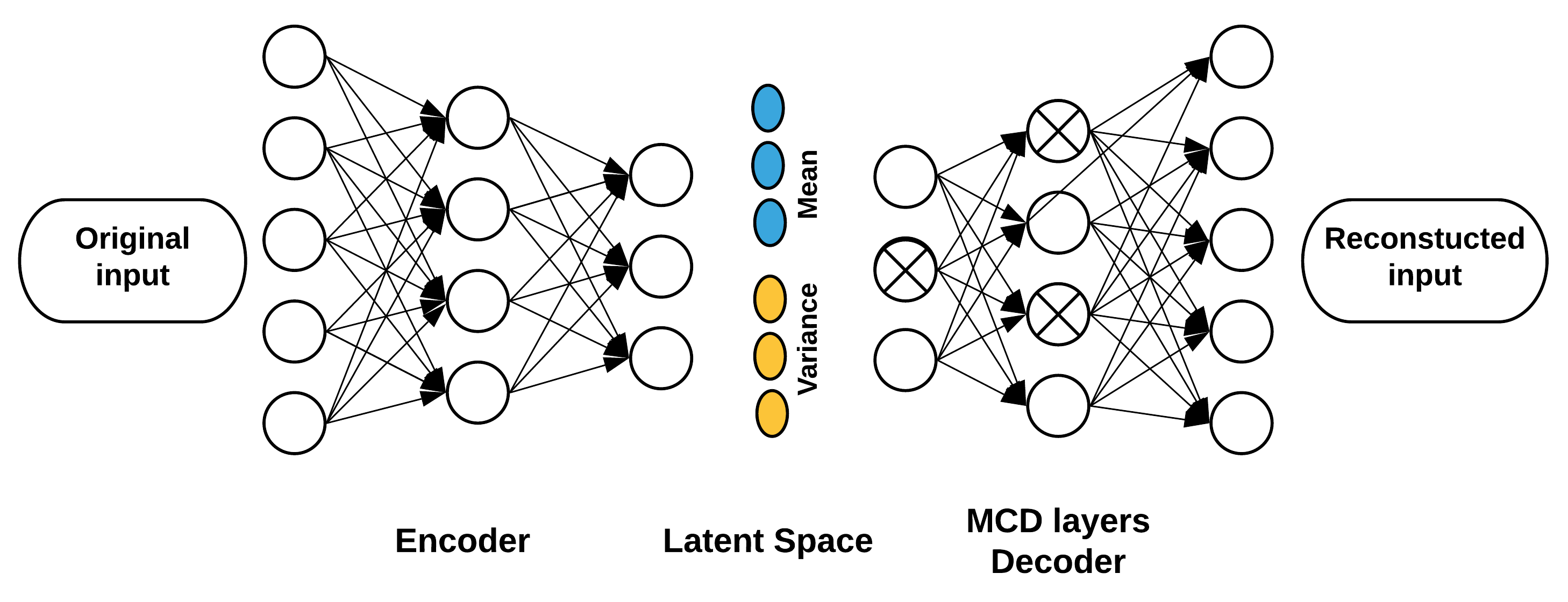}
        \caption{Variational Autoencoder with MCD Decoder. Note the difference to Figure \ref {fig:int}. 
        }
        \label{fig:int2}
\end{figure}

\subsection{VAEs for Data Generation}

For the VAE architecture (Figure \ref{fig:int}), we use two intermediate layers (fully-connected layers) with size of M (e.g. $512$) and N (e.g. $256$) in the encoder. Similarly, the decoder contains two fully-connected layers with N and M neurons. To take into account various types of data sets used in our experiments, we choose the number of latent variables $L$ to be equal to one-half of attributes present in each data set. This value is chosen in order to keep an important part of the information from which the new data can be generated.

There are two approaches to generate the data from the VAE, once the model is trained. The first approach is to generate the sampled latent vectors from the estimated normal distribution $(\mu, \Sigma)$ where the $\Sigma$ is a diagonal covariance matrix. The sampled values are then sent through the decoder part to get the final generated instances. The second approach is to send existing instances through the trained encoder and decoder layers. In this paper, we are interested to generate new values similar to existing values present in the training set, therefore we focus on the second approach.

The process of generating data using the VAE method can be described as follows.
\begin{enumerate}
    \item Obtain the distribution of latent vectors  $(\mu_i, \sigma_i)$ with $i=1, \dots, L$ from each value in the seeding data set by using the encoder.
    \item Resample $t$ times from the obtained latent space distribution, where $t$ is the number of  new instances we want to generate for a single seeding instance as in following equation:
     
     $$
    z_i = \mu_i + \sigma_i \: \cdot \: \epsilon, 
     ~\textrm{where} ~ \epsilon \sim N(0,1).
     $$

    \item Decode the resampled values by the decoder. 
\end{enumerate}

\subsection{MCD-VAE for Data Generation}

The MCD-VAE architecture (Figure \ref{fig:int2}) has a similar structure to the VAE generator, with the exception that the MCD regularization is used within the decoder layers.

The process of generating data with MCD-VAE can be described as follows.

\begin{enumerate}
    \item Obtain the distribution of latent vectors  $(\mu_i, \sigma_i)$ with $i=1, \dots, L$ from each value in the seeding data set.
    \item Send the means $\mu_{1}, \dots, \mu_{L}$ through the MCD decoder $t$ times, where $t$ is the number of  new instances we want to generate for a single seeding instance.
\end{enumerate}

As evident from the above description, MCD-VAE utilizes MCD within the decoder part to get additional fine grained control over the generated instances. Namely, once the 
MCD-VAE is prepared for a single seeding instance, due to dropout, it can produce many different outputs by going forward through the network. This increases the speed of generation and gives the user of the generator much finer control on the generated instances. 

\subsection{MCD-AE for Data Generation}

We can apply the MC dropout method also in the  decoder part of AE and get the generator called  MCD-AE. The structure of MCD-AE in our experiments is similar to VAE and MCD-VAE described before.  The process to generate data in MCD-AE is outlined below.
\begin{enumerate}
    \item For each value in the seeding data set, obtain latent vectors of size $L$.
    \item Send the latent vectors through the MCD decoder. The decoder samples a new dropout mask in each of the $t$ forward passes through the network and generates $t$ values for a single input. 
 \end{enumerate}

The decoder part of the MCD-AE generator is identical to the decoder in MCD-VAE. The difference between the two generators is that MCD-AE does not assume any distributional constraints for the latent space representation.   

\subsection{Visual Comparison of the Generators}
\begin{figure}[t!]
  \centering
  \begin{subfigure}[b]{0.1\linewidth}
    \includegraphics[scale=0.12]{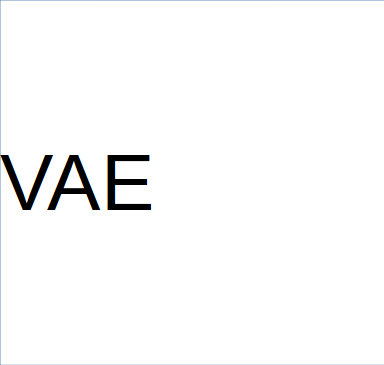}
  \end{subfigure}
  \begin{subfigure}[b]{0.1\linewidth}
    \includegraphics[scale=0.255]{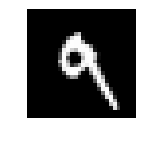}
  \end{subfigure}
  \begin{subfigure}[b]{0.1\linewidth}
    \includegraphics[scale=0.255]{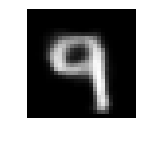}
  \end{subfigure}
  \begin{subfigure}[b]{0.1\linewidth}
    \includegraphics[scale=0.255]{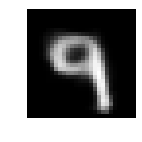}
  \end{subfigure}
  \begin{subfigure}[b]{0.1\linewidth}
    \includegraphics[scale=0.255]{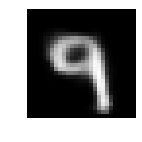}
  \end{subfigure}
  \begin{subfigure}[b]{0.1\linewidth}
    \includegraphics[scale=0.255]{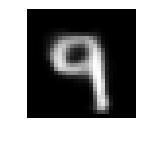}
  \end{subfigure}
  \begin{subfigure}[b]{0.1\linewidth}
    \includegraphics[scale=0.255]{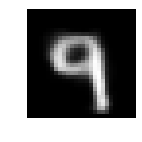}
  \end{subfigure}

  \vspace{-2mm}
  \begin{subfigure}[b]{0.1\linewidth}
    \includegraphics[scale=0.12]{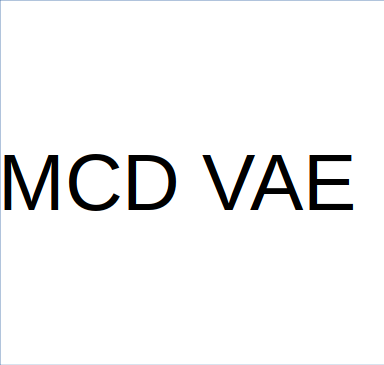}
  \end{subfigure}
  \begin{subfigure}[b]{0.1\linewidth}
    \includegraphics[scale=0.255]{mnist/7_origin.png}
  \end{subfigure}
  \begin{subfigure}[b]{0.1\linewidth}
    \includegraphics[scale=0.255]{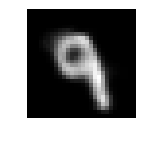}
  \end{subfigure}
  \begin{subfigure}[b]{0.1\linewidth}
    \includegraphics[scale=0.255]{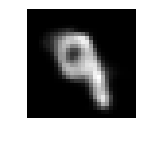}
  \end{subfigure}
  \begin{subfigure}[b]{0.1\linewidth}
    \includegraphics[scale=0.255]{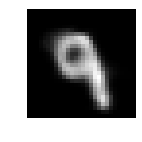}
  \end{subfigure}
  \begin{subfigure}[b]{0.1\linewidth}
    \includegraphics[scale=0.255]{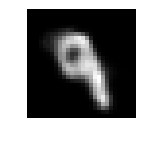}
  \end{subfigure}
  \begin{subfigure}[b]{0.1\linewidth}
    \includegraphics[scale=0.255]{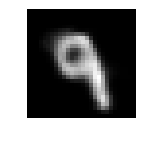}
  \end{subfigure}

  \vspace{-2mm}
  \hspace{5pt}
  \begin{subfigure}[b]{0.1\linewidth}
    \includegraphics[scale=0.11]{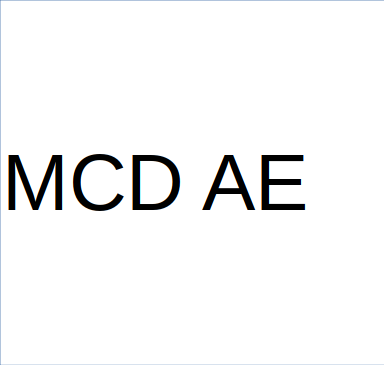}
  \end{subfigure}
  \hspace{4pt}
  \begin{subfigure}[b]{0.1\linewidth}
    \includegraphics[scale=1.0]{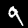}
  \end{subfigure}
  \begin{subfigure}[b]{0.1\linewidth}
    \includegraphics[scale=1.0]{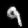}
  \end{subfigure}
  \begin{subfigure}[b]{0.1\linewidth}
    \includegraphics[scale=1.0]{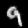}
  \end{subfigure}
  \begin{subfigure}[b]{0.1\linewidth}
    \includegraphics[scale=1.]{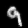}
  \end{subfigure}
  \begin{subfigure}[b]{0.1\linewidth}
    \includegraphics[scale=1.]{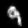}
  \end{subfigure}
  \begin{subfigure}[b]{0.1\linewidth}
    \includegraphics[scale=1.]{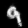}
  \end{subfigure}

\vspace{3mm}
  \hspace{5pt}
  \begin{subfigure}[b]{0.1\linewidth}
    \includegraphics[scale=0.12]{vae_pic.png}
  \end{subfigure}
  \hspace{10pt}
  \begin{subfigure}[b]{0.1\linewidth}
    \includegraphics[scale=0.99]{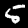}
  \end{subfigure}
  \begin{subfigure}[b]{0.1\linewidth}
    \includegraphics[scale=1.3]{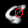}
  \end{subfigure}
  \begin{subfigure}[b]{0.1\linewidth}
    \includegraphics[scale=1.3]{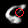}
  \end{subfigure}
  \begin{subfigure}[b]{0.1\linewidth}
    \includegraphics[scale=1.3]{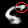}
  \end{subfigure}
  \begin{subfigure}[b]{0.1\linewidth}
    \includegraphics[scale=1.3]{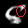}
  \end{subfigure}
  \begin{subfigure}[b]{0.1\linewidth}
    \includegraphics[scale=1.3]{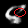}
  \end{subfigure}

 \vspace{-1mm}
  \centering
  \hspace{5pt}
  \begin{subfigure}[b]{0.1\linewidth}
    \includegraphics[scale=0.12]{mcdvae_pic.png}
  \end{subfigure}
  \hspace{10pt}
  \begin{subfigure}[b]{0.1\linewidth}
    \includegraphics[scale=1.0]{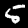}
  \end{subfigure}
  \begin{subfigure}[b]{0.1\linewidth}
    \includegraphics[scale=1.0]{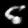}
  \end{subfigure}
  \begin{subfigure}[b]{0.1\linewidth}
    \includegraphics[scale=1.0]{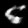}
  \end{subfigure}
  \begin{subfigure}[b]{0.1\linewidth}
    \includegraphics[scale=1.0]{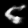}
  \end{subfigure}
  \begin{subfigure}[b]{0.1\linewidth}
    \includegraphics[scale=1.0]{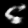}
  \end{subfigure}
  \begin{subfigure}[b]{0.1\linewidth}
    \includegraphics[scale=1.0]{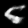}
  \end{subfigure}

 \vspace{-1mm}
  \centering
  \hspace{5pt}
  \begin{subfigure}[b]{0.1\linewidth}
    \includegraphics[scale=0.12]{mcd_ae_pic.png}
  \end{subfigure}
  \hspace{10pt}
  \begin{subfigure}[b]{0.1\linewidth}
    \includegraphics[scale=1.0]{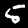}
  \end{subfigure}
  \begin{subfigure}[b]{0.1\linewidth}
    \includegraphics[scale=1.0]{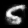}
  \end{subfigure}
  \begin{subfigure}[b]{0.1\linewidth}
    \includegraphics[scale=1.0]{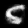}
  \end{subfigure}
  \begin{subfigure}[b]{0.1\linewidth}
    \includegraphics[scale=1.0]{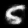}
  \end{subfigure}
  \begin{subfigure}[b]{0.1\linewidth}
    \includegraphics[scale=1.0]{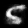}
  \end{subfigure}
  \begin{subfigure}[b]{0.1\linewidth}
    \includegraphics[scale=1.0]{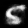}
  \end{subfigure}

 \vspace{3mm}
  \hspace{5pt}
  \begin{subfigure}[b]{0.1\linewidth}
    \includegraphics[scale=0.12]{vae_pic.png}
  \end{subfigure}
  \hspace{5pt}
  \begin{subfigure}[b]{0.1\linewidth}
    \includegraphics[scale=1.0]{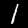}
  \end{subfigure}
  \begin{subfigure}[b]{0.1\linewidth}
    \includegraphics[scale=1.0]{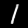}
  \end{subfigure}
  \begin{subfigure}[b]{0.1\linewidth}
    \includegraphics[scale=1.0]{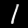}
  \end{subfigure}
  \begin{subfigure}[b]{0.1\linewidth}
    \includegraphics[scale=1.0]{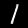}
  \end{subfigure}
  \begin{subfigure}[b]{0.1\linewidth}
    \includegraphics[scale=1.0]{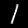}
  \end{subfigure}
  \begin{subfigure}[b]{0.1\linewidth}
    \includegraphics[scale=1.0]{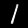}
  \end{subfigure}

\vspace{-1mm}
  \hspace{5pt}
  \begin{subfigure}[b]{0.1\linewidth}
    \includegraphics[scale=0.12]{mcdvae_pic.png}
  \end{subfigure}
  \hspace{5pt}
  \begin{subfigure}[b]{0.1\linewidth}
    \includegraphics[scale=1.0]{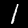}
  \end{subfigure}
  \begin{subfigure}[b]{0.1\linewidth}
    \includegraphics[scale=1.0]{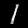}
  \end{subfigure}
  \begin{subfigure}[b]{0.1\linewidth}
    \includegraphics[scale=1.0]{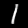}
  \end{subfigure}
  \begin{subfigure}[b]{0.1\linewidth}
    \includegraphics[scale=1.0]{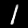}
  \end{subfigure}
  \begin{subfigure}[b]{0.1\linewidth}
    \includegraphics[scale=1.0]{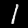}
  \end{subfigure}
  \begin{subfigure}[b]{0.1\linewidth}
    \includegraphics[scale=1.0]{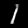}
  \end{subfigure}

\vspace{-1mm}
  \hspace{5pt}
  \begin{subfigure}[b]{0.1\linewidth}
    \includegraphics[scale=0.12]{mcd_ae_pic.png}
  \end{subfigure}
  \hspace{5pt}
  \begin{subfigure}[b]{0.1\linewidth}
    \includegraphics[scale=1.0]{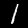}
  \end{subfigure}
  \begin{subfigure}[b]{0.1\linewidth}
    \includegraphics[scale=1.0]{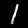}
  \end{subfigure}
  \begin{subfigure}[b]{0.1\linewidth}
    \includegraphics[scale=1.0]{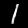}
  \end{subfigure}
  \begin{subfigure}[b]{0.1\linewidth}
    \includegraphics[scale=1.0]{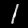}
  \end{subfigure}
  \begin{subfigure}[b]{0.1\linewidth}
    \includegraphics[scale=1.0]{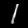}
  \end{subfigure}
  \begin{subfigure}[b]{0.1\linewidth}
    \includegraphics[scale=1.0]{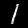}
  \end{subfigure}
    \caption{The generated numbers $9$, $5$, and $1$ are grouped in the top, middle and bottom, respectively.
    Each block of images contains the original seeding image (in the first column) and five generated images using VAE (the first row), MCD VAE (the second row),  MCD AE (the third row). 
    }
    \label{tabe4}
\end{figure}

We visually demonstrate the differences between the three generators (VAE, MCD-VAE, and MCD-AE). For this we have chosen a well-known MNIST data set of hand-written digits\footnote{\url{http://yann.lecun.com/exdb/mnist/}} and used it to train the three generators.  The architecture for VAE and MCD-VAE generators  contains a fully-connected layer with  $1024$ units and a latent layer with the size $10$. The generated images are presented in Figure \ref{tabe4}. The original seeding image is always given in the first column. In this experiment, we investigated generation of digits $9$, $5$, and $1$ (see the three blocks of images). Digits $9$ and $5$ were generated from seeding instances that are written in nonstandard way, with the shape that differs from the rest of digits in their class. The digit $1$ that was used as a seeding instance is written in a standard way.

The five images generated for the digit $9$ using VAE (top group, first row) have the same structures as the seeding digit $9$ but do not reflect much specifics of the seeding image. Contrarily, the images generated using MCD-VAE and MCD-AE (top group, second and third row) tend to better reflect the actual structure of the seeding images.
The digit $5$, used as a seeding instance in the middle group of images is a complete outlier - on the first sight one can not be sure if it is $5$ or $6$. The five generated images for digit $5$ using VAE (middle group, first row) reflect all the training instances and do not take specifics of the seeding instance into account; hence, VAE generates images a bit similar to the digit $8$. On the other hand, the images generated using MCD-VAE and MCD-AE better mimic the seeding image. 
The images generated from the seeding digit $1$, written in the standard way, do not seem to differ much between the three generators (bottom group).

\section{Experimental setting}
In this section, we first describe the methodology used to compare  original and generated data in
Section \ref{sec:experimentalSetting}. We compare  statistical, structural, and prediction properties of two data sets presented in Section \ref{sec:comparisonMethodology}. In Section \ref{sec:datasets}, we present the data sets which served as original data in our evaluation.

\subsection{Data Generation Experiment}
\label{sec:experimentalSetting}
\begin{figure}[hbt]
  \centering
    \includegraphics[width=0.8\linewidth]{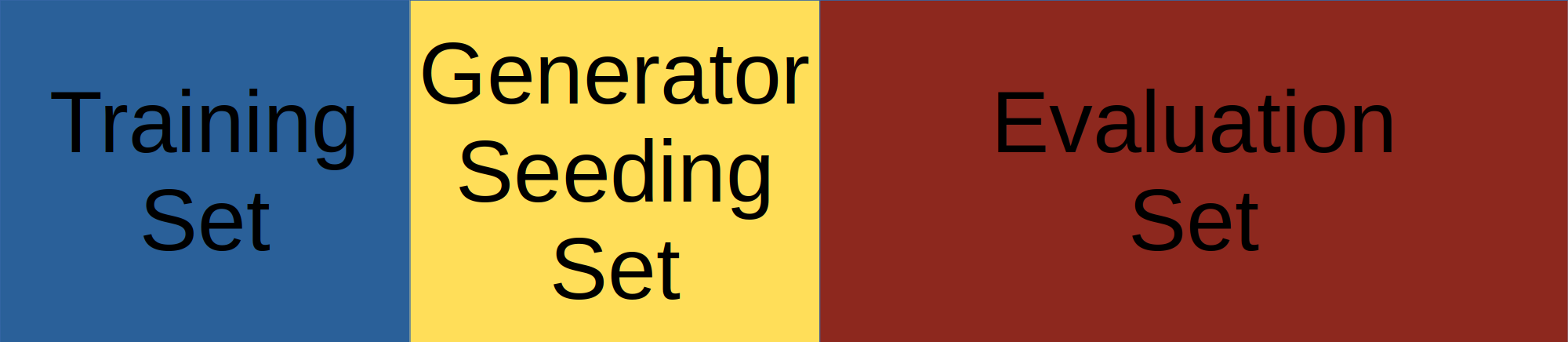}
    \caption{Splits of each original data set used in the experimental evaluation: the generator training set ($25\%$), generator seeding set ($25\%$), and  evaluation set ($50\%$).}
    \label{fig:train_test}
\end{figure}

To prepare a training data set for generators, the original data set is randomly split into two equal parts as shown in Figure \ref{fig:train_test}. The first part is further split into the equal-sized training and generator seeding parts, while the second part of the original data set is left for evaluation. The training part is used to train the generators, while the generator seeding part is used in data generation. From each instance in the generator seeding set, two new instances were generated. Thus, the newly generated data sets are of the same size as the evaluation data sets.

In order to deal with multi-valued categorical attributes, we encode them with several binary substitute attributes, where the presence of a given categorical value in the original attribute sets the substitute variable corresponding to that value to 1.
For example, for a multi-valued attribute $X$ with three values $\{red, green, blue\}$ we form three substitute binary variables $X_{red}, X_{green}, X_{blue}$. If the original attribute contains value $X=blue$, the values of the substitute attributes are  $X_{red}=0, X_{green}=0, X_{blue}=1$. After the data is generated, we perform the reverse operation and decode the substitute variables into one multi-valued attribute.

\subsection{Data Set Comparison}
\label{sec:comparisonMethodology}
In evaluation, presented in Section \ref{sec:evaluation}, we take an existing data set and based on it we generate three synthetic data sets, using VAE, MCD-VAE, and MCD-AE.
The original and the three generated data sets are compared using a general data set comparison framework \cite{robnik2018dataset} which consist of three components, statistical evaluation of differences between attributes, structural comparison of data sets based on clustering, and predictive comparison based on classification models. We describe the three components below.
\begin{enumerate}
    \item \textit{Statistical evaluation of attributes.} test the mean, standard deviation, and differences in distributions between matching attributes in two compared data sets. In order to make comparison sensible for all statistics, the attributes are normalized to $[0,1]$ scale. The value that summarizes the difference between the two data sets is calculated as the median value of pairwise attribute differences. For example, to compare mean across the whole data set, we compute the differences in means for each of the attributes and then average these values and report it as the final measure. We therefore report $\Delta$mean and $\Delta$std. 
    
    \item \textit{Clustering performance evaluation} is performed based on the structured based distance comparing two data sets using the adjusted Rand index (ARI) \cite{hubert1985comparing}. The ARI value is in range of $[0,1]$ having 0 in the case of random distributions of clusters and 1 for ideally matching clusters. The clusters of two data sets are separately computed and the process obtains the medoids for each of the clusterings. The instances in the second data set are assigned to the nearest clusters in the first data set based on the medoids computed for the first data set. The same assignment is repeated with the first data set, as instances of the first data set are assigned to clusters computed on the second data set based on the medoids from these clusters. In this way, we obtain two clusterings that contain instances from both data sets. Finally, we use ARI to summarize the clustering similarity between the two clusterings and report it as the data sets topological similarity value.   
    
    \item \textit{Classification performance} based evaluation measures the predictive similarity of two data sets by comparing random forest classification accuracies on the two data sets. Let us assume that the original data set is denoted as $d_1$ and the generated data sets are labeled with $d_2$. Both $d_1$ and $d_2$ are split into two parts, where the first parts are used to train the random forest models $m_1$ and $m_2$, while the second parts are used for testing. Four accuracy values are computed: $m_1d_1$ - model computed on the first data set and evaluated on the first data set; $m_1d_2$ - model computed on the first data set and evaluated on the second data set; $m_2d_1$ - model computed on the second data set and evaluated on the first data set; and $m_2d_2$ - model computed on the second data set and evaluated on the second data set. If those four values are similar (in particular if accuracies on the original data set are close, i.e. accuracies of $m_1d_1$ and $m_2d_1$), one can conclude that the first and the second data set have similar predictive performance. We report only the difference of $m_2d_1 -m_1d_1$ as the predictive similarity $\Delta$acc.
    \end{enumerate}

\subsection{Data Sets}
\label{sec:datasets}
To evaluate the difference between results of the three generators,  we use data sets from UCI (University of California Irvine) repository \cite{bache2013uci}. The R package readMLDATA \cite{savicky2012readmldata} was used for data manipulation. We selected  classification data sets with between 500 and 1000 instances. The characteristics of the used data sets are provided in Table \ref{tab2}. 

\begin{table}[H]
\footnotesize
	\centering
	\caption{\footnotesize The characteristics of the used data sets. The columns are: $n$ - number of instances, $a$ - number of attributes, num - number of numeric attributes, disc - number of discrete attributes, v/a - average number of values per discrete attribute, C - number of class values, majority \% - proportion of majority class in percentages, missing \% - percentage of missing values. 
	}
	\renewcommand{\arraystretch}{1}
	\setlength{\tabcolsep}{2.5pt}
\begin{tabular}{lrrccrrrr}
 & & & & & & & \textbf{majority} & \textbf{missing} \\
	\textbf{Data set}& \textbf{n} & \textbf{a} & \textbf{num} & \textbf{disc} & \textbf{v/a} & \textbf{C} & \textbf{(\%)} & \textbf{(\%)} \\	

		\hline		
\textbf{Brest-WDBC}  & 569 & 30  & 30  & 0  & 0.0  & 2  & 62.7 & 0.00 \\
\textbf{Brest-WISC}  & 699 & 9  & 9  & 0  & 0.0  & 2  & 65.5 & 0.25 \\
\textbf{Credit-screening}  & 690 & 15 &  6 & 9 & 4.4 & 2 & 55.5 & 0.64 \\
\textbf{PIMA-diabetes}  & 768 & 8 &  8 & 0 & 0.0 & 2 & 65.1 & 0.00 \\
\textbf{Statlog-German} & 1000 & 20  & 7  & 13  & 4.2  & 2  &  70.0 & 0.00  \\
\textbf{Tic-tac-toe}  & 958 & 9  & 0  & 9  & 3.0  & 2  &  65.3 & 0.00  \\

		\hline
	\end{tabular}
	
	\label{tab2}
\end{table}

\section{Evaluation and results}
\label{sec:evaluation}
Using the above described data sets we evaluated the quality of data generators. 
In Table \ref{tab01} we compare the original data set with the generated data set using VAE architecture. The results comparing the original data set with the MCD-VAE and MCD-AE generators are presented in Tables \ref{tab02} and \ref{tab03}, respectively. For comparison we use the statistical, structural, and predictive criteria, described in Section \ref{sec:comparisonMethodology}, i.e. the average difference in means ($\Delta$mean) and standard deviation ($\Delta$std), similarity of produced clusters expressed with Adjusted Rand Index (ARI), and differences in predictive accuracy $ \Delta $acc ($m_2d_1 -m_1d_1$). 

\begin{table}[H]
\footnotesize
	\centering
	\caption{\footnotesize Comparison between the original data and VAE generator.}
	\renewcommand{\arraystretch}{1.1}
	\setlength{\tabcolsep}{3.5pt}
\begin{tabular}{lrrrr}
	\textbf{Data set}& \textbf{$\Delta$ mean} & \textbf{$\Delta$ std} & \textbf{ARI} & \textbf{$\Delta$acc} \\

		\hline		

\textbf{Breast-WDBC}  & -0.161 & -0.089  &  0.909 &  -0.024 \\
\textbf{Breast-WISC}  & -0.069 & 0.001   & 0.970 & -0.045 \\
\textbf{Credit-screening} & -0.078 &  -0.041  & 0.474 & -0.068 \\
\textbf{PIMA-diabetes} &  -0.171 &  -0.047  & 0.446 &  -0.015 \\

\textbf{Statlog-German} & -0.040  &  0.040  &  0.167 & -0.000 \\

\textbf{Tic-tac-toe}  & - & -   & 0.133 & -0.092 \\



		\hline
	\end{tabular}
	
	\label{tab01}
\end{table}

\begin{table}[H]
\footnotesize
	\centering
	\caption{\footnotesize Comparison between the original data and MCD-VAE generator. 
	}
	\renewcommand{\arraystretch}{1.1}
	\setlength{\tabcolsep}{3.5pt}
\begin{tabular}{lrrrr}
	\textbf{Data set}& \textbf{$\Delta$ mean} & \textbf{$\Delta$ std} & \textbf{ARI} & \textbf{$\Delta$acc} \\ 

		\hline		

\textbf{Breast-WDBC}  & -0.045 & -0.044  & 0.876 & -0.008\\
\textbf{Breast-WISC}  &  0.011 & 0.013  & 0.916 &  -0.011 \\

\textbf{Credit-screening} &  -0.028 &  -0.038   & 0.447 & -0.061 \\

\textbf{PIMA-diabetes} & -0.022 &  -0.028  & 0.715 & -0.007 \\

\textbf{Statlog-German} &  -0.016 & 0.028  &  0.243 & -0.001  \\

\textbf{Tic-tac-toe}  & - & - & 0.122 & -0.017 \\



		\hline
	\end{tabular}
	
	\label{tab02}
\end{table}

\begin{table}[H]
\footnotesize
	\centering
	\caption{\footnotesize Comparison between the original data and MCD-AE generator. 
	}
	\renewcommand{\arraystretch}{1.1}
	\setlength{\tabcolsep}{3.5pt}
\begin{tabular}{lrrrr}
	\textbf{Data set}& \textbf{$\Delta$ mean} & \textbf{$\Delta$ std} & \textbf{ARI} & \textbf{$\Delta$acc} \\ 

		\hline		

\textbf{Breast-WDBC}  &-0.059 & -0.048  & 0.746 & -0.014\\
\textbf{Breast-WISC}  &  0.004 &  0.021  & 0.994 &   -0.020\\

\textbf{Credit-screening} & -0.046 &  -0.036  & 0.393 & -0.072 \\

\textbf{PIMA-diabetes} &  -0.077 &  -0.037  & 0.551 & -0.012 \\

\textbf{Statlog-German} &  -0.030 & 0.021  &  0.235 & 0.000  \\

\textbf{Tic-tac-toe}  & - & - & 0.224 & -0.158 \\



		\hline
	\end{tabular}
	
	\label{tab03}
\end{table}

Comparing the results in Tables \ref{tab01}, \ref{tab02}, and \ref{tab03}, we can see that differences between the original and generated data are small. There is no clear pattern which of the three generators is better. We can conclude that all of them are useful, while minor differences in the quality of the generated data may depend on the structure of a data set. However, it can be observed that MCD-VAE provide slightly better classification performance than VAE and MCD-AE based on the compared $\Delta$acc value. On the other hand, for \textit{Breast-WDBC}, \textit{Breast-WISC} and \textit{Credit-screening} datasets VAE generator has the better clustering performance than the two newly introduced generators.


\section{Comparing efficiency of generators}
In order to compare the data generation time (in seconds) of VAE, MCD-VAE, and MCD-AE, we measure the time for $100$ repetitions of the data generating process using the above described data sets. To get reliable measurements, we resample each seeding instance $1000$ times (instead of $2$ times as in the previous experiments). Table \ref{tab:comparision} reports the mean and standard deviation of the measured times. 
We generate data sets as described in Section \ref{sec:methods}
For VAE, the instances in seeding data sets are encoded to obtain the latent values, then the latent values are resampled and decoded. For MCD-VAE and MCD-AE, we obtain the mean values with the seeding instances and obtain the generated data using the MCD decoder. 

\begin{table}[H]
\footnotesize
	\centering
	\caption{\footnotesize Comparison of time required for data generation in seconds. }
	\renewcommand{\arraystretch}{1.1}
	\setlength{\tabcolsep}{3.5pt}
\begin{tabular}{lrrrrrr}
	\textbf{Datasets/Models}& \textbf{VAE [s.d.]} & \textbf{MCD-VAE [s.d.]} & \textbf{MCD-AE [s.d.]}\\
		\hline		
\textbf{Breast-WDBC}  & 1.04 [0.018] &  0.89 [0.022] & 0.89 [0.020] \\
\textbf{Breast-WISC}  & 0.90 [0.019] & 0.85 [0.037] & 0.89 [0.011] \\
\textbf{Credit-screening} & 1.00 [0.030] & 0.93 [0.025] & 0.94 [0.010]\\
\textbf{PIMA-diabetes} & 0.91 [0.034] & 0.85 [0.021] & 0.85 [0.016] \\
\textbf{Statlog-German} & 1.07 [0.018] & 1.03 [0.018] & 1.10 [0.045] \\
\textbf{Tic-tac-toe}  & 0.99 [0.026] &  0.93 [0.039] & 0.94 [0.012]\\
		\hline
	\end{tabular}
	\label{tab:comparision}
\end{table}

The MCD-VAE and MCD-AE generators are consistently slightly faster than the VAE generator (between 5-10\%). Although the MCD-AE generator is  architecturally simpler, it is not faster then the MCD-VAE generator. The datasets used are relatively small, hence, for the larger datasets, we expect larger differences. 

\section{Conclusions and Further Work}
We constructed and compared three generators of semi-artificial data. The VAE generator is based on the variational autoencoder architecture while the  MCD-AE and MCD-VAE employ Monte Carlo dropout within autoencoders and variational autoencoders. The comparison of the generated data sets based on statistical, structural, and predictive properties shows that the three generators produce similar data sets which are highly similar to the original data.

The advantages of the proposed Monte Carlo dropout employed within VAE and AE over the existing VAE method can be summarized with the following two points:
\begin{itemize}
    \item Improved speed. Based on the results presented in Table \ref{tab:comparision} we can conclude that generating data using MCD-VAE and MCD-AE is slightly faster than using the VAE generator.
    \vspace{0.1cm}
    \item Greater flexibility. The MCD-VAE and MCD-AE methods generates data similar to specific selected seeding instances. This can be very useful if the provided seeding instances are outliers or instances of special interest. For example, in image generation, the newly generated images will be closer to the original one even when the original image is different from the rest of the images in the training set.   
\end{itemize}

The advantage of the MCD-AE method over MCD-VAE method is that does not make any distributional assumptions during the latent space generation. The information received from the encoder part is directly introduced into the MCD decoder. The time required for data generation using MCD-AE is similar to MCD-VAE. The more detailed differences between these generators are left for further investigation. 

With methodological development of deep learning, the models that can estimate the distributions, e.g., the variational autoencoders, are becoming increasingly important. Hence, our further work will focus on investigating new and improving existing architectures that can generate new data efficiently and reliably. Further, we aim to test those architectures within different application contexts. As bio-medical imaging is expensive and limited by the budget, our goal is to investigate data generation within this field.  
The Python code of the proposed generators is publicly available\footnote{\url{https://github.com/KristianMiok/MCD-VAE}}. 

\subsection*{Acknowledgement}
The work was supported by the Slovenian Research Agency (ARRS) core research programme P6-0411 (Marko Robnik-\v{S}ikonja). The research was carried out in the frame of the project Bioeconomic approach to antimicrobial agents - use and resistance financed by UEFISCDI by contract no. 7PCCDI / 2018, cod PN-III-P1-1.2-PCCDI-2017-0361 (Kristian Miok and Daniela Zaharie). This project has also received funding from the European Union’s Horizon 2020 research and innovation programme under grant agreement No 825153 (EMBEDDIA) (Kristian Miok and Marko Robnik-\v{S}ikonja).
\bibliographystyle{ICCP2019}
\bibliography{ICCP2019}
\end{document}